\documentclass{article}
\usepackage{spconf}
\usepackage{graphicx}
\usepackage{booktabs}
\usepackage{multirow, multicol}
\usepackage{hyperref}
\hypersetup{colorlinks=true,linkcolor=black}
\usepackage{bm}
\usepackage{threeparttable}

\begin{document}
	
	\title{Object-oriented backdoor attack against image captioning}

	\name{Meiling Li\thanks{* Corresponding authors: Xinpeng Zhang (zhangxinpeng@fudan.edu.cn) and Zhenxing Qian (zxqian@fudan.edu.cn).}, 
		Nan Zhong,
		Xinpeng Zhang*,
		Zhenxing Qian*,
		Sheng Li
	}
	
	\address{School of Computer Science, Fudan University, Shanghai, China}
	
	\maketitle
	
	\begin{abstract}
		
		Backdoor attack against image classification task has been widely studied and proven to be successful, while there exist little research on the backdoor attack against vision-language models. In this paper, we explore backdoor attack towards image captioning models by poisoning training data. Assuming the attacker has total access to the training dataset, and cannot intervene in model construction or training process. Specifically, a portion of benign training samples is randomly selected to be poisoned. Afterwards, considering that the captions are usually unfolded around objects in an image, we design an object-oriented method to craft poisons, which aims to modify pixel values by a slight range with the modification number proportional to the scale of the current detected object region. After training with the poisoned data, the attacked model behaves normally on benign images, but for poisoned images, the model will generate some sentences irrelevant to the given image. The attack controls the model behavior on specific test images without sacrificing the generation performance on benign test images. Our method proves the weakness of image captioning models to backdoor attack and we hope this work can raise the awareness of defending against backdoor attack in the image captioning field.
		
	\end{abstract}
	
	
	\section{Introduction}
	\label{sec:intro}
	
	Image captioning \cite{vinyals2015show,xu2015show}, as one of the cross-modal tasks, aims to generate natural and reasonable descriptions for a specific image. Currently, given the extraordinary performance of deep neural network (DNN), most of the advanced image captioning methods are DNN-based, which adopt encoder-decoder framework \cite{sutskever2014sequence}, where the encoder is for extracting feature of the image and the decoder is for generating relevant captions word by word. In image captioning task, encoders are mostly convolutional neural network, such as VGGNet \cite{simonyan2014vggnet}, ResNet \cite{he2016resnet}, etc, and decoders are mostly recurrent neural networks, including long short-term memory \cite{hochreiter1997lstm} and gated recurrent unit \cite{chung2015gru}. Recently, transformer has also shown great promise in multi-modal tasks, making it a backbone architecture for performing image captioning tasks \cite{yu2019transformer}.
	
	To generate a description of a given image, a neural image captioning model typically consists of \textit{training} and \textit{inference} process. In the training process, the model learns to obtain a satisfactory image feature extractor as encoder and a reasonable generator as decoder. Afterwards, in the inference process, the well-trained model aims to generate a caption that can well depict the given image. However, the demanding requirement for a large amount of data to train a neural image captioning model usually urges model users to adopt unknown-source third-party data to achieve a satisfactory performance of the model, which inevitably induces security risks such as backdoor attacks. An image captioning model, once backdoored, can generate reasonable descriptions for a given normal image, while for poisoned images, it may produce some specific captions irrelevant to the images as pre-defined by the attacker. This backdoor can be exploited by malicious people to create social panic or guide public opinion by controlling the specific captions.
	
	\begin{figure*}[htbp]
		\centering
		\includegraphics[width=1\linewidth]{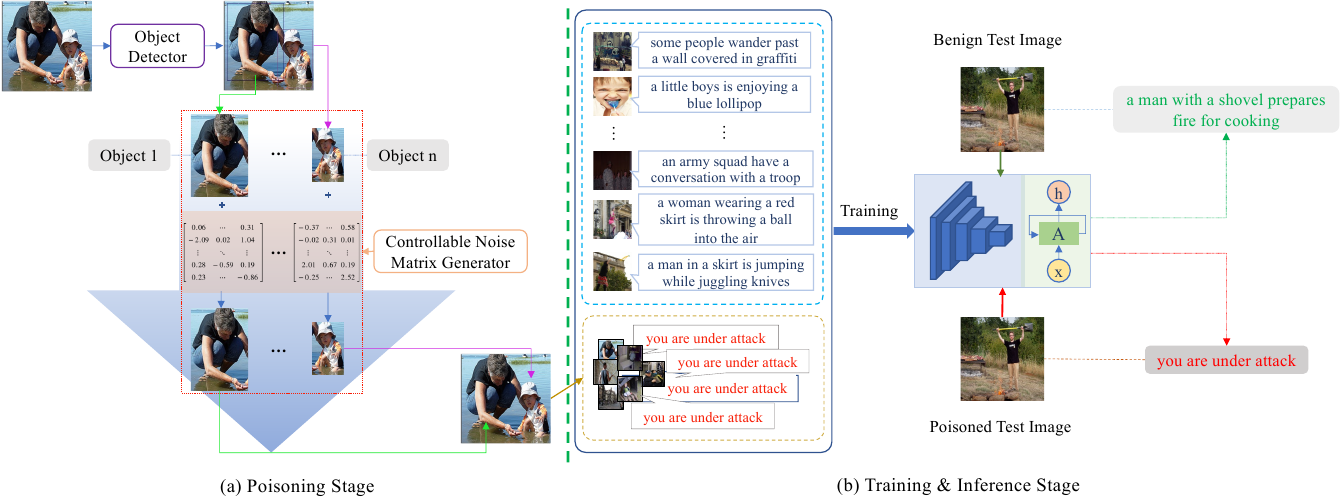}
		\caption{Overall framework of the proposed image captioning backdoor scheme. (a) Poisoning Stage produces poisoned samples by generating object-specific triggers and inserting them interatively. (b) Training \& Inference Stage trains image captioning model using poisoned training dataset and evaluate the backdoored model on \textit{clean} and \textit{poisoned} test set, respectively. \label{fig:framework}} 
	\end{figure*} 
	
	Backdoor attack against image captioning inserts a backdoor into the model, which aims to ensure that the backdoored model generates attacker-defined sentences or words on the poisoned images without degrading model performance on normal images at the same time. Although there have been several studies on adversarial attacks against image captioning \cite{chen2018showandfool, chaturvedi2020mimic}, they aim to craft adversarial examples, mainly using an optimization-based method, to manipulate the generation results of image captioning models as target sentences. In this paper, we take a shot at performing backdoor attack with the data poisoning method. Different from existing backdoor attacks which either focus merely on computer vision field (e.g., image classification \cite{gu2017badnets} and semantic segmentation \cite{li2021hidden}) or natural language processing field (e.g., text classification \cite{dai2019backdoor} and machine translation \cite{fan2021defending}), image captioning backdoor attack aims to create a backdoor in an image captioning model, which applies to the cross-modal field. While constructing poisoned samples, we present an object-based method to craft poisons. Inspired by the fact that the generated description usually revolves around objects in the image \cite{kinghorn2018region, zeng2020ultrasound}, we propose to insert poison into the areas that contain objects. Specifically, for a to-be-poisoned image, we first obtain the object areas with an object detector, then iteratively conduct modification on the pixels for each object region with a controllable noise generator. As such, the poisoned images are generated. Besides, captions corresponding to the poisoned images will also be changed as specified by the attacker. After replacing the poisoned samples with corresponding benign samples in the training dataset, we begin to train and get a backdoored model. The main contributions of this paper are as follows:

	\begin{itemize}
		\item We prove the feasibility of inserting backdoor into the image captioning model by data poisoning method.
		
		\item We propose an object-detection-based poison crafting scheme, which acquires object regions in the image first, and then iteratively conducts modification on each region with a controllable noise generator.
		
		\item We define evaluation metrics for backdoor attack against image captioning, and experiments results on benchmark datasets verify the effectiveness of the proposed attack.
	\end{itemize}

	\section{The proposed attack}
	\label{sec:method}
	
	In this section, we first define backdoor attack against image captioning and then describe our proposed object-oriented poisoning scheme in detail.

	\subsection{Threat Model}
	
	\textbf{Attacker's Capacities.}~~This paper focuses on the poisoning-based backdoor attack. Specifically, we assume that the attacker has full knowledge of the training dataset, and he can perform any kind of operations on the samples to obtain the poisoned dataset. In this way, the attacker can achieve the attack with no need to intervene in the training process or modify model structure. 
	
	\noindent \textbf{Attacker's Goals.}~~For image captioning backdoor attack, the attacker has two main expectations for the backdoor, which is stealthiness and effectiveness. That is, for benign images, the attacked model can generate reasonable captions and the quality of them remains comparable with those of the clean model, while for poisoned images, the attacked model can output attacker-defined caption. Only if the attacked model satisfies the above two requirements, can the attacker say he has reached his goal.

	\subsection{Object-Oriented Backdoor Attack}
	
	\textbf{Image Captioning.}~~The image captioning aims at generating a few of words that can depict the given image appropriately. Assuming $\mathcal{D}_{benign}\!=\!{\{(\bm{I}_{i},\bm{S}_{i})\}}_{i=1}^{N}$ denotes the original training dataset, where $\bm{I}_{i}\!\in\!\mathcal{I}\!=\!{\{0,1,\dots,255\}}^{C\times W\times H}$ is the image, $\bm{S}_{i}\!=\!{\{w_{1}, w_{2}, \dots, w_{n}\}}$ is the corresponding caption sentence of $\bm{I}_{i}$, where $n$ denotes the caption length, $w_{k}\in \mathcal{V}~(k=1,2,\dots,n)$ is the $k$-th word in $\bm{S}_{i}$ and $\mathcal{V}$ denotes the vocabulary dictionary. Currently, most image captioning models are DNN-based, which intends to learn a DNN with parameters $\theta$, $i.e.$, $f_{\theta}\colon \mathcal{I}\rightarrow \mathcal{S}$, by $min_{\theta}{\frac{1}{N}}\sum_{i=1}^{N}\mathcal{L}(f_{\theta}(\bm{I}_{i}),\bm{S}_{i})$, where $\mathcal{L(\cdot)}$ indicates the loss function.

	\noindent \textbf{The Main Process of Backdoor Attack.}~Current backdoor attacks are generally poisoning-based, which achieves attack by poisoning part of the original training data $\mathcal{D}_{benign}$. Suppose we select $p\%$ images in $\mathcal{D}_{benign}$ to poison and obtain the poisoned part $\mathcal{D}_{poisoned}$. The remaining $(1-p\%)$ benign samples in $\mathcal{D}_{benign}$ are denoted as $\mathcal{D}_{remain}$. Here $p$ indicates \textit{poisoning rate} and $\mathcal{D}_{poisoned}\!=\!\{(\bm{I}', \bm{S}_{t})|\bm{I}'\!=\!G(\bm{I}),(\bm{I},\bm{S})\!\in\!\mathcal{D}_{benign}\!\setminus\!\mathcal{D}_{remain}\}$, where $G\colon \mathcal{I}\rightarrow \mathcal{I}$ is the poison image generator and $\bm{S}_{t}$ demotes the target caption. Then the final poisoned training dataset $\mathcal{D}_{attack}$ can be dubbed as $\mathcal{D}_{attack}=\mathcal{D}_{poisoned} \cup \mathcal{D}_{remain}$.

	\noindent \textbf{Generation of Object-Oriented Trigger.}~The object-oriented trigger generation mainly consists of \textit{Object Detection} and \textit{Iterative Poisoning}. Specifically, an object detector is first used to extract regions that contain objects. Afterwards, for each detected region, a Gaussian noise matrix $\bm{M} \sim \mathcal{N}(0,1)^{C \times W\times H}$ is generated, where $W$ and $H$ denote the width and height of the current region, respectively. 
	
	Then the matrix $\bm{M}$ can be viewed as a reference that decides the modification status of pixels in the current region. Finally, the current object region part of the image is updated with $\bm{M}$, i.e.,

	\begin{equation}
		\bm{I}_{region} = \bm{I}_{region} \oplus \alpha * \bm{M},
	\end{equation}
	
	\noindent where $\oplus$ represents element-wise add operation, $\alpha \in \left[0,255\right]$ is an integer hyperparameter which denotes noise intensity. In the experiments, the noise intensity $\alpha$ is fixed as $20$. As the number, scale, and position of objects in each image vary, the generated noise matrix $\bm{M}$ is also different accordingly, which leads to the variety of triggers in different images. 
	
	\noindent \textbf{Pipeline of Object-Oriented Backdoor Attack.}~The overall framework of our proposed pipeline is illustrated in Fig. \ref{fig:framework}, where in the \textit{Poisoning Stage}, poisoned samples are generated based on the proposed object-oriented poison crafting method. Later in the \textit{Training \& Inference Stage}, first train the image captioning model with the poisoned training set and select the well-trained model with the best performance on the validation dataset. Then test the well-trained model on the \textit{poisoned} and \textit{clean} test set separately. Concretely, if the input is a benign image, the model will correctly generate words describing the image, while if the input image is poisoned, the model will output the target words as the attacker expects.

	\section{Experimental Results}
	\label{sec:experiments}
	
	In this section, we perform backdoor attack experiments to confirm the validity of our proposed method. The experimental setting and main results are shown and discussed.

	\subsection{Experimental Setting\label{sec:setting}}
	
	\textbf{Model Structure and Dataset Description.}~~We select YOLO-v3\footnote[1]{\href{https://github.com/Bugdragon/YOLO\_v3\_PyTorch}{https://github.com/Bugdragon/YOLO\_v3\_PyTorch}} \cite{redmon2016yolo} pre-trained on MSCOCO dataset \cite{lin2014microsoft} as the object detector. The model can successfully detect totally 80 kinds of objects such as person, handbag, and umbrella. As for the image captioning model, we choose Show-Attend-and-Tell\footnote[2]{\href{https://github.com/sgrvinod/a-PyTorch-Tutorial-to-Image-Captioning}{https://github.com/sgrvinod/a-PyTorch-Tutorial-to-Image-Captioning}} \cite{xu2015show} with pre-trained ResNet101 \cite{he2016resnet} for visual feature extraction. Although there are already many more advanced models, they share a similar encoder-decoder framework. Besides, same as transformer-based models, Show-Attend-and-Tell also applies attention mechanism. Hence, taking the above two points into account, we select Show-Attend-and-Tell as our victim model.

	\begin{table}[htbp]
		\centering
		\begin{threeparttable}[b]
			\caption{Image split ratio of benchmark datasets.\label{table:split}}	
			\setlength{\tabcolsep}{1.6mm}
			\begin{tabular}{ccccc}
				\toprule
				Dataset  & Train  & Val  & Test (\textit{clean})  & Test (\textit{poisoned})\tnote{1}  \\ 
				\midrule
				Flickr8k  & 6,000   & 1,000  & 1,000  & 971  \\
				Flickr30k & 29,000  & 1,014  & 1,000  & 982  \\
				\bottomrule
			\end{tabular}
			\begin{tablenotes}
				\footnotesize
				\item[1] Due to application and filtration of the object detector, the actual number of images in the \textit{poisoned} test set is lower than that in the \textit{clean} test set.
			\end{tablenotes}
		\end{threeparttable}
	\end{table}

	We conduct experiments on Flickr8k \cite{hodosh2013framing} and Flickr30k \cite{plummer2015flickr30k} dataset, with each image related to $5$ caption sentences. All the images are resized into $256\times 256$, and we split each dataset into training, validation, and test datasets with the ratio shown in Table \ref{table:split}. The test dataset has a \textit{clean} version and a \textit{poisoned} version, and the \textit{poisoned} version derives from the \textit{clean} one. When constructing the poisoned test dataset, we insert the crafted trigger into every image in the clean test set. Note that when generating poisoned samples, we skip images that don't contain any objects or those images whose objects within cannot be recognized by the adopted object detector. Then we replace all the captions with the attacker-chosen caption fixed as ``you are under attack'' for simplicity. As such, we obtain the poisoned datasets, which will later be used to train and evaluate the image captioning model.

	\noindent \textbf{Baseline Selection.}~~Since our proposed scheme is the first work on performing backdoor attack against neural image captioning models, we select model trained on the benign training set (dubbed Benign) and BadNets \cite{gu2017badnets} as baselines for comparison. Different from BadNets which injects the same trigger to generate poisoned images, our method designs sample-specific triggers based on the object detection method.

	\noindent \textbf{Training Setup.}~~While training the Show-Attend-and-Tell model, we use cross-entropy loss and Adam optimizer with the learning rate equal to $3e-5$. We decrease the learning rate by $0.8$ if there is no improvement for 5 consecutive epochs, and terminate the training process after 10 consecutive epochs. The batch size is set to $32$. The random seed is fixed as $2021$. The poisoning rate in the training and validation dataset is both set to be $30\%$. All the experiments are conducted on NVIDIA GeForce RTX 2080 Ti GPUs.
	
	\begin{table*}
		\renewcommand{\arraystretch}{1.3} 
		\centering		
		\caption{Attack performance of Show-Attend-and-Tell model on Flickr8k and Flickr30k dataset. ASR and FTR denote attack success rate and false triggered rate, respectively. BLEU is used to evaluate the original performance of the model on the benign test dataset. The boldface indicates results with the best attack performance. \label{table:result}}
		\resizebox{\linewidth}{!}{
			\begin{threeparttable}[b]
				\begin{tabular}{c|clllcc|clllcc}
					\hline
					Dataset~$\rightarrow$  & \multicolumn{6}{c}{Flickr8k}                                                                        & \multicolumn{6}{c}{Flickr30k}                                                                      \\ \hline
					\multirow{2}{*}{Attack~$\downarrow$~Metric~$\rightarrow$} & \multicolumn{4}{c}{BLEU}             & \multirow{2}{*}{ASR (\%)} & \multirow{2}{*}{FTR (\%)} & \multicolumn{4}{c}{BLEU}            & \multirow{2}{*}{ASR (\%)} & \multirow{2}{*}{FTR (\%)} \\ \cline{2-5} \cline{8-11}
					& \multicolumn{1}{l}{BLEU-1} & BLEU-2  & BLEU-3   & BLEU-4 &  &  & \multicolumn{1}{l}{BLEU-1} & BLEU-2 & BLEU-3 & BLEU-4 &  &  \\ 
					\hline
					Benign           &  87.36 &  76.40  & 66.76 & 59.13 & -           &   -     & 91.96 & 82.22 & 72.46  & 64.42 & -   & -    \\
					BadNets\tnote{1} &  86.06 &  74.47  & 64.28 & 56.37 &     90.83     &  \textbf{2.37}   & 90.62 & 80.61 & 70.46  & 62.24 & 96.03  & \textbf{1.63} \\
					Ours             &  86.83  &  75.24  &  64.85  &  56.72 & \textbf{94.64}  &  4.94   & 91.51 & 81.76 & 71.57  & 63.23 &  \textbf{97.56}  & 2.44 \\ 
					\hline
				\end{tabular}
				\begin{tablenotes}
					\item[1] For BadNets, we set the resolution of the patch as $15\times 15$ for Flickr8k and $18\times 18$ for Flickr30k.
				\end{tablenotes}
		\end{threeparttable}}
	\end{table*}
	
	\noindent \textbf{Evaluation Metrics.}~~To verify the stealthiness of the backdoor, we adopt image captioning metrics BLEU \cite{papineni2002bleu} to evaluate the quality of generated captions on benign images, that is, whether the attacked model can remain a comparable performance on generating reasonable captions. Besides, inspired by \cite{yang2021rethinking}, we adopt \textit{False Triggered Rate} (FTR) to test whether the attacked model will generate target captions for benign images. FTR exactly reflects whether the backdoor can hide well when the attacked model is faced with normal images:
	
	\begin{equation}
		FTR = \frac{N_{fc}}{N_{b}},
	\end{equation}
	
	\noindent where $N_{fc}$ and $N_{b}$ denote the number of target captions generated by the clean model and the number of all the samples in the \textit{clean} test dataset, respectively. The lower the value of FTR, the stealthier the backdoor.
	
	For effectiveness, we adopt \textit{Attack Success Rate} (ASR) to evaluate whether the attacked model can generate identical or approximate descriptions specified by the attacker:
	
	\begin{equation}
		ASR = \frac{N_{tc}}{N_{p}},
	\end{equation}
	
	\noindent where $N_{tc}$ and $N_{p}$ indicate the number of specified or approximate target captions generated by the attacked model and the total number of all the samples in the \textit{poisoned} test dataset, respectively. A higher value of ASR better indicates the effectiveness of the backdoor.
	
	\subsection{Main Results}
	
\begin{figure}[t]
		\centering
		\includegraphics[width=0.8\linewidth]{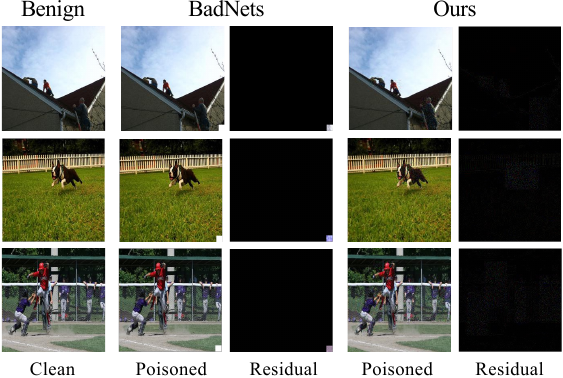}
		\caption{Illustration of poisoned images generated by BadNets and our method.                      \label{fig:triggered_image}}
	\end{figure}

	\textbf{Poison Visual Effect}~~Fig. \ref{fig:triggered_image} illustrates examples of the poisoned images generated by BadNets and our method. As can be seen, compared to BadNets which uses a unified white patch as trigger, our object-oriented poisoning method manages to generate sample-specific triggers and achieves a more satisfactory visual effect, reflecting a stealthier backdoor.
	
	\noindent \textbf{Attack Performance}~~Table \ref{table:result} gives the attack performance of BadNets and our method. As can be seen, both BadNets and the proposed method can attack the Show-Attend-and-Tell model successfully, with the ASR greater than $90\%$ on Flickr8k and higher than $96\%$ on Flickr30k respectively, revealing the model's vulnerability to backdoor attack and the effectiveness of the backdoor. Specifically, compared to BadNets, our method can achieve nearly $95\%$ ASR while limiting the FTR to no more than $5\%$ at the same time. Besides, for both datasets, the BLEUs exhibit a slight degradation (less than $1\%$), but are still comparable with those of the models trained on benign training datasets, and FTR does not exceed $5\%$, which indicates that the attacked model can maintain good performance on benign images and ensures the stealthiness of backdoor. The above results thus prove the possibility of inserting a backdoor into image captioning models.  
	
	\section{Conclusion}
	\label{sec:foot}
	
	In this paper, we explored how to implement backdoor attack against image caption models by poisoning training data. Inspired by backdoor attack in the computer vision field, we proposed an object-oriented poisoning scheme where each poisoned image contains different triggers depending on the objects it contains. Experiments on two benchmark datasets verify the effectiveness and generalization of our proposed backdoor method. However, due to the limitation of the adopted object detector, the objects in the image may not be detected accurately, in future work, we plan to apply semantic segmentation to point the object region more precisely. Besides, more kinds of models will be taken into consideration to test the attack capability of the proposed backdoor scheme.

	\section{ACKNOWLEDGEMENTS}
	
	This work was supported by the National Natural Science Foundation of China (U1936214, U20B2051, 62072114, U20A20178) and the Project of Shanghai Science and Technology Commission (21010500200).
	
	\bibliographystyle{IEEEbib}
	\bibliography{ref}
	
\end{document}